%% file: ICAPS24.tex
\documentclass[letterpaper]{article} % DO NOT CHANGE THIS
\usepackage{aaai24}  % DO NOT CHANGE THIS
\usepackage{times}  % DO NOT CHANGE THIS
\usepackage{helvet}  % DO NOT CHANGE THIS
\usepackage{courier}  % DO NOT CHANGE THIS
\usepackage[hyphens]{url}  % DO NOT CHANGE THIS
\usepackage{graphicx} % DO NOT CHANGE THIS
\usepackage{natbib}  % DO NOT CHANGE THIS AND DO NOT ADD ANY OPTIONS TO IT
\usepackage{caption} % DO NOT CHANGE THIS AND DO NOT ADD ANY OPTIONS TO IT
\usepackage{lipsum}
\usepackage{amsmath}
\usepackage{amssymb}
\usepackage[ruled, vlined, linesnumbered]{algorithm2e}
\SetKw{Continue}{continue}
\SetKw{And}{\textbf{and}}
\SetKw{Or}{\textbf{or}}
\SetKw{In}{\textbf{in}}
\SetKw{To}{\textbf{to}}
\usepackage{etoolbox}
\makeatletter
\newcommand{\nosemic}{\renewcommand{\@endalgocfline}{\relax}}% Drop semi-colon ;
\newcommand{\dosemic}{\renewcommand{\@endalgocfline}{\algocf@endline}}% Reinstate semi-colon ;
\let\oldnl\nl% Store \nl in \oldnl
\newcommand{\nonl}{\renewcommand{\nl}{\let\nl\oldnl}}% Remove line number for one line
\patchcmd{\@algocf@start}% <cmd>
  {-1.5em}% <search>
  {0pt}% <replace>
  {}{}% <success><failure>
\makeatother

\usepackage{siunitx}

\usepackage{color}
\usepackage{subcaption}
\usepackage{rotating}
\usepackage{multirow}
\usepackage{arydshln}
\usepackage{makecell}
\usepackage[switch, modulo]{lineno}
\title{Combined Task and Motion Planning Via Sketch Decompositions \\ (Extended Version with Supplementary Material)}
\author{%
Magí Dalmau-Moreno$^{1,2}$, 
Néstor García$^1$, 
Vicenç Gómez$^2$ and  
Héctor Geffner$^{3,4}$
}
\affiliations{
$^1$Eurecat, Centre Tecnològic de Catalunya, Spain\\
$^2$Universitat Pompeu Fabra, Barcelona, Spain\\
$^3$RWTH Aachen University, Germany\\
$^4$Link{\"o}ping University, Link{\"o}ping, Sweden \\
\{magi.dalmau, nestor.garcia\}@eurecat.org,
vicen.gomez@upf.edu, hector.geffner@ml.rwth-aachen.de
}

\input{commands}

\newcommand{\vicenc}[1]{{\color{magenta}{\textbf{Remark Vicen\c{c}:} #1}}}
\newcommand{\hector}[1]{{\color{red}{\textbf{Remark Hector:} #1}}}
\newcommand{\magi}[1]{{\color{blue}{\textbf{Remark Magi:} #1}}}

\begin{document}

\maketitle

\begin{abstract}
The challenge in combined task and motion planning (TAMP) is the effective integration of a search over a combinatorial space, usually carried out by a task planner, and a search over a continuous configuration space, carried out by a motion planner.
Using motion planners for testing the feasibility of task plans and filling out the details is not effective because it makes the geometrical constraints play a passive role. This work introduces a new interleaved approach for integrating the two dimensions of TAMP that makes use of \emph{sketches}, a recent simple but powerful language for expressing the decomposition of problems into subproblems.
A sketch has width~$1$ if it decomposes the problem into subproblems that can be solved greedily in linear time. In the paper, a general sketch
is introduced  for several classes of TAMP problems which has width~$1$  under suitable assumptions.
While sketch decompositions have been developed for classical planning, they offer two important benefits in the context of TAMP.
First, when a task plan is found to be unfeasible due to the  geometric constraints, the combinatorial search resumes in a  specific subproblem. Second, the sampling of object configurations is not done once, globally, at the start of the search, but locally, at the start   of each  subproblem.
%Finally, the sketches provide  a convenient interface for integrating the functionality of the task and motion planners  as they split  the problem into subproblems, some of which are to be solved by the former and some by the latter. 
Optimizations of this basic setting are also considered  and  experimental results over existing and new pick-and-place benchmarks are reported.
% , including a Blocks-World domain in a 3D-spatial setting.
\end{abstract}

\input{sections/introduction.tex}

\input{sections/tasks.tex}
%\input{sections/background.tex}
\input{sections/cmtp_with_sketches.tex}

\input{sections/related_work}

\input{sections/experimental_results.tex}

\input{sections/conclusion}

\input{sections/acknowledgments}

\bibliography{bibliography.bib}
\newpage
\input{sections/background}

\input{sections/supplementary-pddlstream}

\end{document}

%% file: commands.tex
\newcommand{\tup}[1]{(#1)}

\newcommand{\iw}[1]{\ensuremath{\text{IW}(#1)}\xspace}

\newcommand{\siw}{\ensuremath{\text{SIW}}\xspace}
\newcommand{\siwR}{\ensuremath{\text{SIW}_{R}\xspace}}

\newcommand{\Q}{\mathcal{Q}}

\newcommand{\pplus}{\hspace{-.05em}\raisebox{.15ex}{\footnotesize$\uparrow$}}
\newcommand{\mminus}{\hspace{-.05em}\raisebox{.15ex}{\footnotesize$\downarrow$}}

\newcommand{\DEC}[1]{#1\mminus}
\newcommand{\INC}[1]{#1\pplus}
\newcommand{\UNK}[1]{#1?}

\newcommand{\Omit}[1]{}

\newcommand{\prule}[2]{\{ #1 \} \mapsto \{ #2 \}}

\newcommand{\domain}{\ensuremath{D}}
\newcommand{\instance}{\ensuremath{I}}
\newcommand{\goal}{\ensuremath{\mathit{Goal}}}
\newcommand{\initial}{\ensuremath{\mathit{Init}}}

\newcommand{\states}{\ensuremath{S}}
\newcommand{\initialstate}{\ensuremath{s_0}}
\newcommand{\goalstates}{\ensuremath{G}}
\newcommand{\actions}{\ensuremath{\mathit{Act}}}
\newcommand{\successor}{\ensuremath{f}}
\newcommand{\applicability}{\ensuremath{A}}

%% file: sections/introduction.tex
% \magi{

% General comments:
% \begin{itemize}
   
% \end{itemize}
% }
%  {\color{red} *** Changed $n$ and $s$ for $u$ and $v$ respectively, so as not to confuse with state $s$ and path length $n$}
\section{Introduction}
\begin{figure*}[t!]
    \centering
    \includegraphics[height=2.9cm]{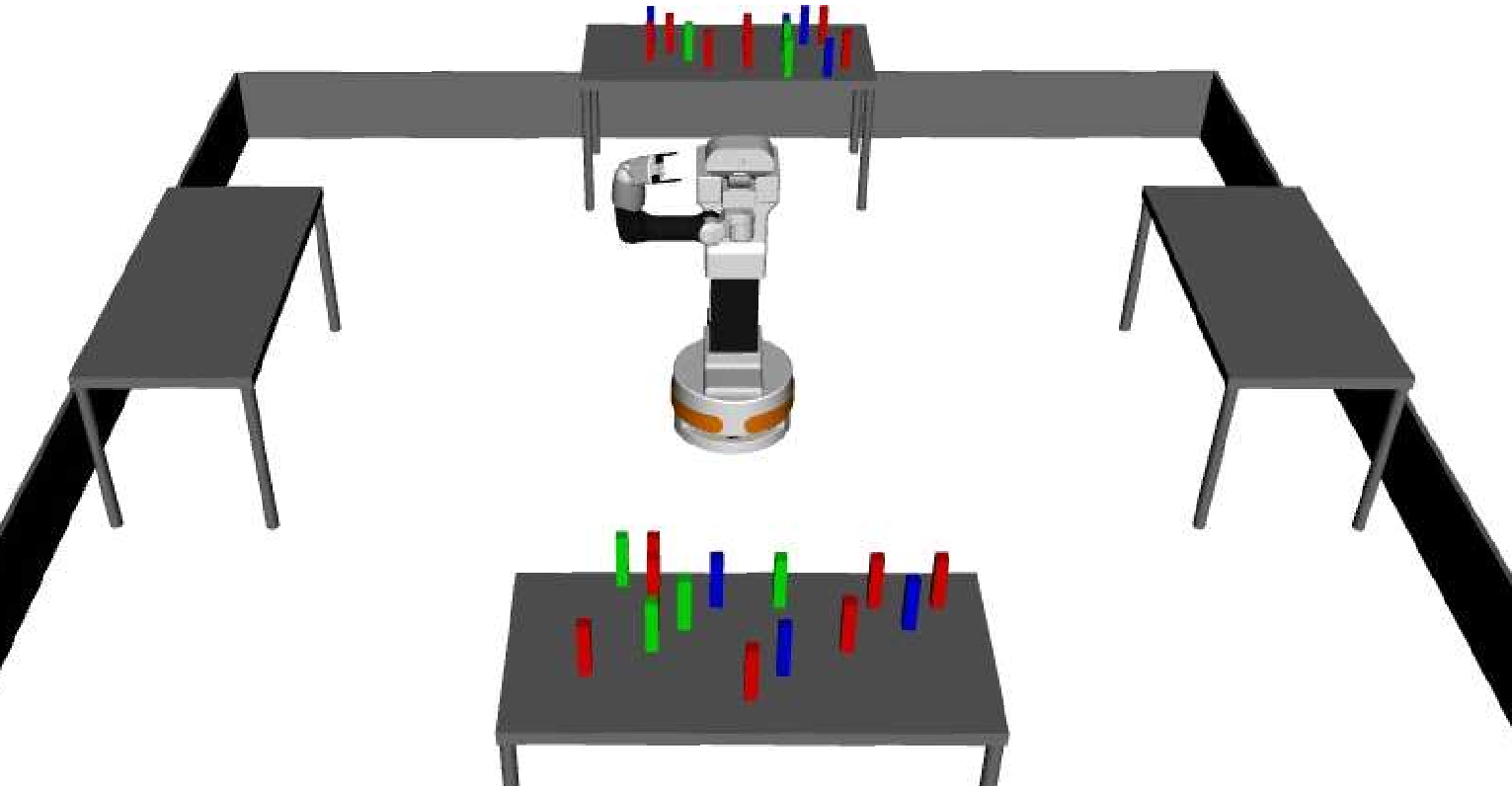}\hfill
    \includegraphics[height=2.9cm]{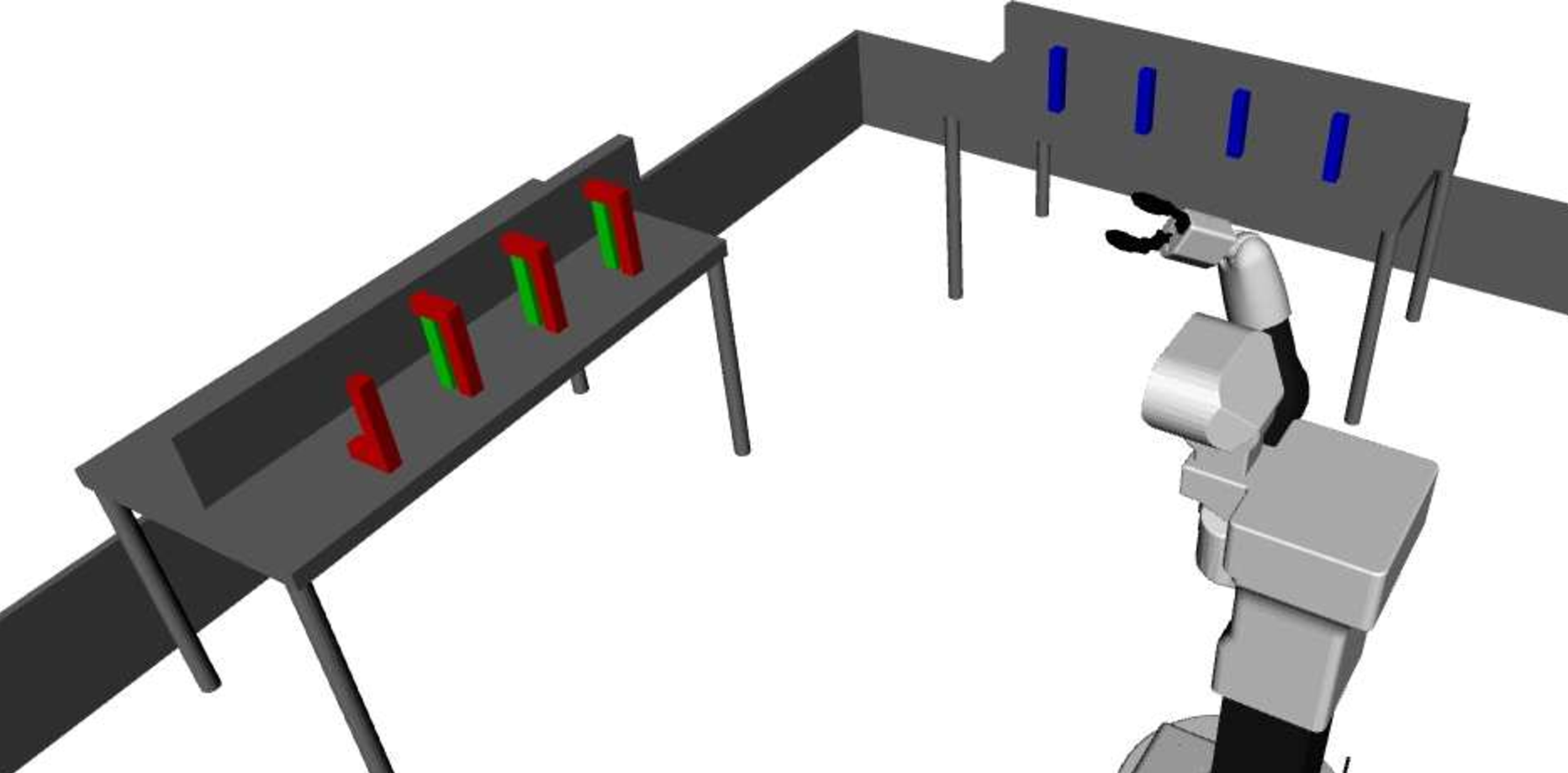}\hfill
    \includegraphics[height=2.9cm]{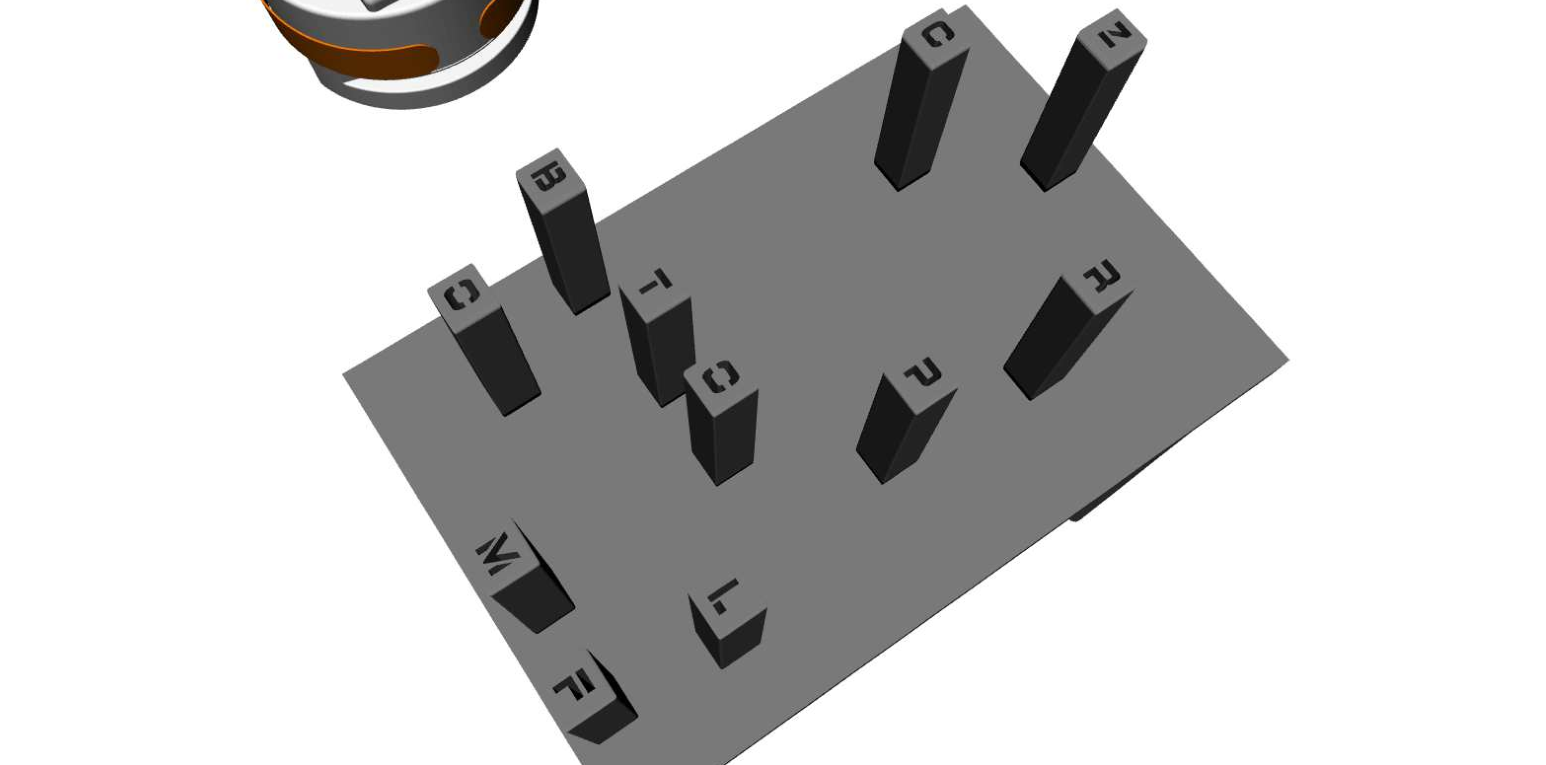}
    \caption{Examples of TAMP considered in this work: \emph{Sorting Objects} (left),
    \emph{Non-Monotonic} (middle), and 
    {\emph{Words}} (right).}
    \label{fig:problems}
\end{figure*}

Combined task and motion planning \cite{
doi:10.1146/annurev-control-091420-084139,6943079}%cambon_hybrid_2009,10.1007/11008941_11,10.3389/frobt.2021.637888
, refers to planning problems where a robot manipulates  objects in an environment in order to achieve a given goal (Figure~\ref{fig:problems}).
The main challenge in TAMP  is to integrate planning at two levels effectively:  1)~task planning, which requires searching over a combinatorial space to find a sequence of ``high-level'' symbolic actions; 
and 2)~motion planning, which requires searching for ``low-level'' paths through the robot's continuous state space.
There are two basic approaches for TAMP~\cite{doi:10.1146/annurev-control-091420-084139}: 1)~to search for high-level plans which are subsequently checked for geometric and kinematic feasibility using a low-level solver; and 2)~to iteratively generate, typically by sampling, feasible configurations (robot poses, grasps, etc.) which are then used to generate a global sequence of high-level actions.
% In both cases, the integration between the two planning levels is poor, as the geometrical constraints play a passive role in the search for plans.

% \begin{figure}[t!]
%     \centering
%     \includegraphics[width=\columnwidth]{Figures/sorting_objects}\vspace{1mm}
%     \includegraphics[width=\columnwidth]{Figures/non_monotonic}\vspace{1mm}
%     \includegraphics[width=\columnwidth]{Figures/blocksworld2}\vspace{1mm}
%     %\includegraphics[width=0.9\columnwidth]{Figures/blocks_world} % 
%     \caption{Examples of TAMP problems considered in this work: \emph{Sorting Objects} (top),
%     \emph{Non-Monotonic} (middle), and 
%     {\emph{Words}} (bottom).}
%     \label{fig:problems}
% \end{figure}

In this work, we present a novel interleaved TAMP approach to integrate  task  and  motion planning that takes  advantage of
the  powerful language of \emph{sketches} developed recently in  classical planning for decomposing problems into subproblems~\cite{bonet-geffner-aaai2021} which  can then be solved by means of effective width-based search algorithms
\cite{Lipovetzky2021WidthBasedAF}%lipovetzky-geffner-ecai2012,
. Sketches are collections of rules over state
features that can be crafted by hand, expressing domain knowledge \cite{drexler-et-al-kr2021}, or can be learned automatically \cite{drexler2022learning}.
% Sketches  are high-level rules that allow to express domain-specific knowledge in terms of subproblems.
% Sketches enable width-based search algorithms~\cite{lipovetzky-geffner-ecai2012} to solve complex domains effectively, in provable polynomial time.
% The use of sketches facilitates a more active role of the geometrical constraints during the plan search.
% This has two important consequences.
The use of sketches for  TAMP  has two advantages: 1)~when a task plan is found to be unfeasible due to geometric constraints, the combinatorial search resumes in a specific subproblem; and 2)~the sampling of configurations is performed locally, at the start of each  subproblem instead of globally, at preprocessing. 
While sketches and width-based search have been developed for  classical planning, neither one requires
declarative action models in PDDL or the like, and only require suitable state features to be defined over
the states obtained from a simulator. When the  subproblems have a width bounded by $k$, a simple \siwR~procedure
solves the problems in time exponential in~$k$~\cite{bonet-geffner-aaai2021}. For our TAMP tasks, we craft a sketch
that yields subproblems of  width 1 under suitable assumptions that are discussed. 

% This work requires several assumptions regarding the computation of sketch features and the sampling of configurations to proof probabilistic completeness. We characterize theoretically such assumptions and show experimentally that they are satisfied in practical pick-and-place benchmarks.

The paper is organized as follows. We describe first the tasks,  review the notions of sketches and width,  present the TAMP  formulation
and the sketches for dealing with three families of pick-and-place tasks, the experimental results, related work, and conclusions.

%% file: sections/tasks.tex
\section{Benchmarking TAMP tasks}\label{sec:tasks}

All the tasks considered as benchmarks in this work, depicted in Figure~\ref{fig:problems}, require object manipulations through pick-and-place actions in cluttered scenarios and have the same high-level goal of ensuring that no object remains misplaced, which is a goal that fits many robotic manipulation tasks. 
The first two tasks are taken from a TAMP benchmark~\cite{Lagriffoul2018PlatformIndependentBF} that aims to be a standard in this field.
The third task is inspired by the classical \emph{Blocks World}.

\emph{Sorting Objects---}
    A robot must arrange different blocks standing on different tables, based on their color.
    The goal constraints are that all $N$ blue blocks must be on the left table and all $N$~green blocks must be on the right table. There are also $2N$~red blocks, acting as obstacles for reaching blue and green blocks, whose goal position is free.
    The robot is allowed to freely navigate around the tables, while keeping within the arena, picking and placing the blocks at the tables.
    The proximity between the blocks forces the planner to carefully order the operations, as well as to move red blocks out of the way without creating new obstructions, i.e.,~blocking objects.
    Thus, this problem requires to move many objects, sometimes multiple times. %, i.e.~large task-space.

\emph{Non-Monotonic---}
    A robot must move three green blocks standing on a table to their corresponding goal position on another table.
    At the initial state, there are four red blocks obstructing the direct grasp of the green blocks and there are also four blue blocks obstructing the direct placement of the green blocks at their target position.
    In any case, the red and blue blocks must end up being at the same initial locations.
    The robot is allowed to freely navigate around the tables, while keeping within the arena, picking and placing the blocks at the tables.
    The goal condition on blue and red blocks requires to temporarily move them away and bring them back later on (non-monotonicity), to solve the task.

\emph{Words---}
A robot must arrange anywhere on a table some blocks, each one labelled with a letter (possibly repeated), to build a target word, e.g.,~{\tt TAMP} and {\tt ROBOT}.
Then, the goal positions of each block are not absolute but relative to the positions of other blocks, and blocks can share goal positions if a letter is repeated in the goal sequence.
Besides, the available free space on the table is limited, and thus there are obstructions between blocks.
This modification is intended to make the coupling of the geometric and symbolic reasoning more challenging.  
Thus, a block is well-placed if, in a given state, it belongs to the largest valid sequence (by \emph{valid} we mean that the sequence is within the goal sequence and located far enough away from the edges of the table to allow the whole sequence to be completed). 

% \hector{
% HOW we'll do it  (sketch, etc).

% Less than 1 page; perhaps 1 colum + figs. 
% }

%% file: sections/cmtp_with_sketches.tex
\section{Combined Task and Motion Planning}

We assume the reader is familiarized with the framework of width-based planning and sketches. A background section, providing a review of these concepts, is available in the supplementary material.

\begin{figure*}[t]
    \centering
    \begin{subfigure}[b]{0.196\textwidth}
        \centering
        \includegraphics[width=\textwidth]{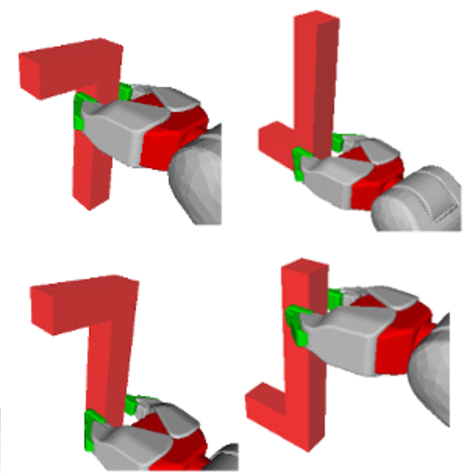}
        
    \caption{}
        \label{fig:grasps}
    \end{subfigure}
    \hfill
    \begin{subfigure}[b]{0.196\textwidth}
        \centering
        \includegraphics[width=\textwidth]{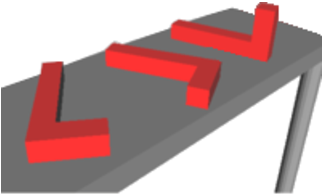}
        
    \caption{}
        \label{fig:sops}
    \end{subfigure}
    \hfill
    \begin{subfigure}[b]{0.196\textwidth}
        \centering
        \includegraphics[width=\textwidth]{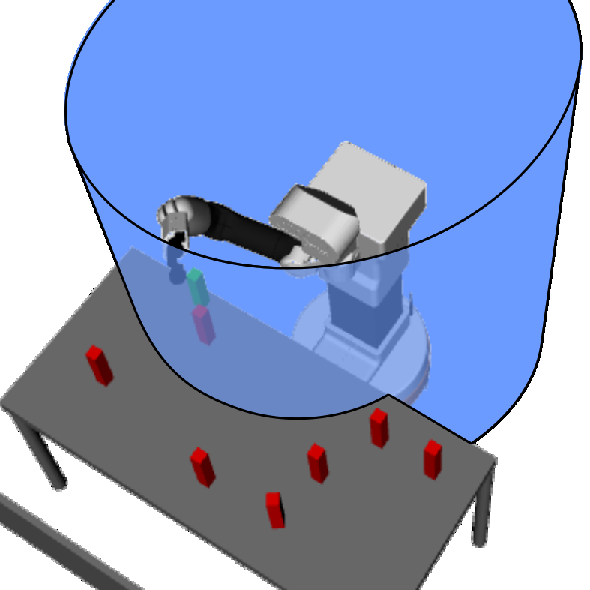}
        
    \caption{}
        \label{fig:in_workspace}
    \end{subfigure}
    \hfill
    \begin{subfigure}[b]{0.196\textwidth}
        \centering
        \includegraphics[width=\textwidth]{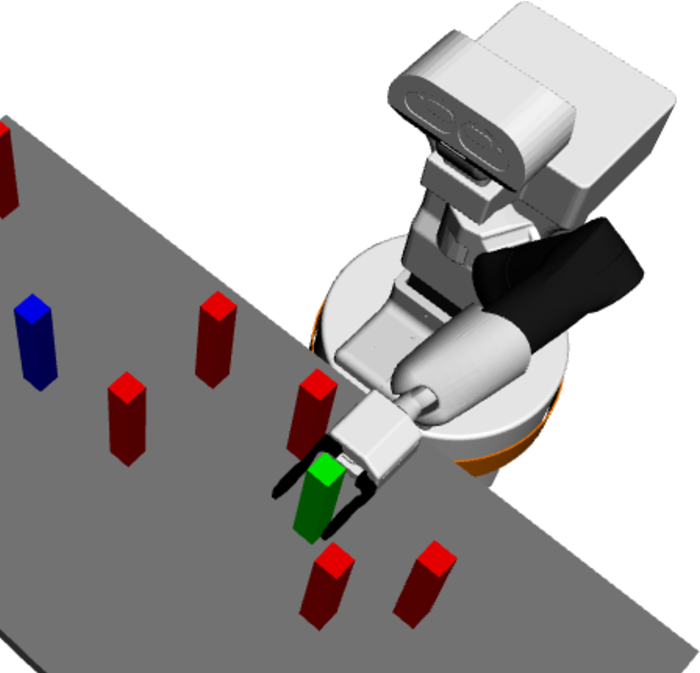}
        
    \caption{}
        \label{fig:inverse_kinematics}
    \end{subfigure}
    \hfill
    \begin{subfigure}[b]{0.196\textwidth}
        \centering
        \includegraphics[height=\textwidth]{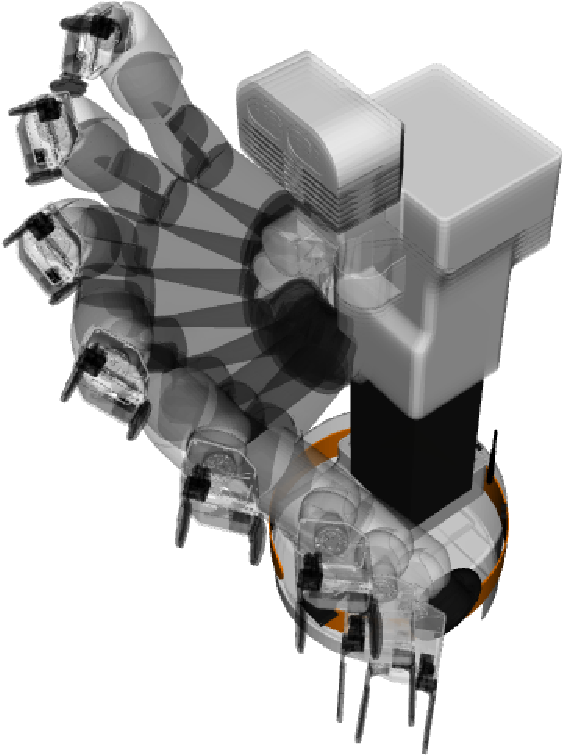}
        
    \caption{}
        \label{fig:motion_plan}
    \end{subfigure}%
    \caption{Different grasps~(a), SOPs~(b), and calls to \emph{InArmWorkspace}~(c), \emph{InverseKinematics}~(d), and \emph{MotionPlan}~(e) functions.}
    \label{fig:actions}
\end{figure*}

\Omit{
\vicenc{
Describe formally the TAMP problem in general (the what, not the how)
\begin{itemize}
\item similar from what described previously but for $S$ being the configuration (continuous) space
\item also introduce here the actions, which operate on a continuous-valued state
\item mention the constraints: applicability conditions and the motion constraints 
\item goal as a set of configuration states
\item define what is a plan in this case as a solution to the problem
\item after this section the reader could verify whether a plan solves the problem or not
\end{itemize}
}
}

We consider pick-and-place problems where, at a high-level, a mobile robot with an arm can perform three actions: picking up an object from a surface,
placing it on a surface, and moving to another place (dragging any grasped object).
Underlying this action-space there is a complex, lower-level problem involving a continuous state-space (position values of robot joints, robot and object locations and spatial constraints).
% robot configuration, object poses, and spatial constraints.
The formulation exploits a particular width-based task-planner based on
the {\siwR} algorithm~\cite{bonet-geffner-aaai2021} guided by a simple and general sketch~$R$ with a handful of chosen state-features that decompose the problems into subproblems solved by a linear-time search guided by IW(1)~\cite{lipovetzky-geffner-ecai2012}.
This task-planner calls a motion planner for checking the feasibility of high-level actions and for driving the joints.
Two characteristics of the proposed integration is that the sampling required for mapping the high-level action schemas into ground actions is done at the level of subproblems, and that ``backtracking'', namely failure of high-level subplans, is carried out also at the subproblem level.
There is no ``deep backtracking'', the {\siwR}  search proceeds forward to the goal, one subgoal at a time by means of the linear-time IW(1) procedure, and it is the width~$w$ of the sketch over the family~$\Q$ of pick-and-place tasks that ensures that this search will be complete; namely, when $w=1$. 

\subsection{Problem Formulation}

The class of TAMP tasks that we address can be characterized as a tuple $T = \tup{S,s_0,G,Act,A,X,N}$ where:

\begin{itemize}
\item $S$ represents the continuous states with all relevant  information, $s_0$ the initial state and $G$ the set of goal states;
\item $Act$ is the set of high-level action schemas;
\item $A$ is the \emph{sampling function} that given a state $s$ and the set of action schemas $Act$ produces the set of \emph{sampled grounded actions} $A(s)$ that are relevant in $s$ (although not necessarily executable in $s$).
Thus, $A(s)$ limits the set of states $S$ that the planner can potentially visit given $s$;
\item $X$ is an \emph{executability function} that maps ground actions $a \in A(s)$ and 
states $s \in S$ into \emph{motion plans}  $X(a,s)=\rho$ that implement the action $a$ in $s$ at the low level, if the ground action is feasible; if not, $X(a,s)=\bot$;
\item $N$ is the state-transition function that yields the next state~$s'$ when the motion plan $X(a,s)=\rho \not= \bot$, is executed in $s$;
i.e.~$s' = N(s,\rho)$ when $\rho \not=\bot$. 
\end{itemize}

A \emph{plan} or \emph{solution} for a tuple $T$ is a sequence of ground actions $a_0,\ldots,a_n$ such that:
1)~$a_{i}\!\in\!A(s_i)$,  \mbox{$i=0,\ldots,n$};
2)~\mbox{$s_{i+1} \!=\! N(s_i,\rho_i)$} for \mbox{$\rho_i\!=\!X(a_i,s_i)$}, \mbox{$\rho_i \!\not=\! \bot$}, \mbox{$i=0,\ldots, n$}; and
3)~$s_{n+1} \in G$.
%% Thus, only actions sampled in a feasible state can be chosen.
%% Hector: Confusing as stated. Removed
Starting with~$s_0$, the state progresses according to low-level motion plans and the
given dynamics, and this  progress must end in a goal state. 

% This formulation calls for modeling TAMP tasks in a certain way, by proving all the components of $T$, and for a task planner to compute the action sequences that solve~$T$.
%% Hector. Confusing as stated. Removed.
In the search, the main bottleneck is in the execution function~$X(a,s)$ that is expensive as it involves run-time calls to
a motion planner.
Note also that one special case of the above formulation arises when the configuration sampling that leads to the finite set~$A(s)$ of ground actions is done only once at pre-processing.
In such a case, $A(s)=A(s_0)$ for all states.
Other variations do one sampling per subproblem; in such a cases $s$ ``borrows'' the set of ground actions from other states $s'$ considered before as $A(s)=A(s')$.

% {\color{red}
% ** Below: express things below specification of the components of $T$ above, in order; try to be precise/exact **}

\subsection{State-Space $S$, start state $s_0$ and goal states $G$}
% {\color{red}
% **** Describe: what is the state that the motion planner sees; it's only continous; e.g, how feature $H$ is evaluated from state? Is $H$ represented also in low level state? If so, then state space is hybrid: continuous and discrete **

% **** This is too vague .... be crisp, precise
% }
% \nestor{motion planner sees only continuous states: e.g. $H$ can be inferred by checking that the object center of mass is located within the gripper fingers}
% \hector{Existing text, not mine}

On the one hand, the state space $S$ is continuous and comprises the robot configuration-space (i.e.~the position value of the robot joints), the robot pose (i.e.~position and orientation in Cartesian space) together with the pose of the movable objects.
Formally, $S \in \mathbb{R}^d \times \text{SE}(3) \times \dots \times \text{SE}(3),$
where~$d$ is the number of degrees of freedom of the robot
%, $O$ is the number of movable objects 
and SE$(3)$ is the special Euclidean group of homogeneous transformations, one for the robot and as many as objects.
Note that $S$ is what the motion-planner ``sees'' and that low-level discrete information can be inferred from it (e.g.~knowing that an object is grasped by checking that its position coordinates are within the gripper fingers).
On the other hand, the initial state $s_0$ involves the starting configuration of the robot and the objects and the set~\goalstates\ is the set of states~$s$ that imply that all misplaced objects are in their goal positions/regions.

%In the implementation of this work, we consider a robot with nine degrees of freedom (seven for the arm, one for the torso height, and one for the gripper openness) and poses are expressed by tridimensional position vectors and orientation quaternions.
%However, the proposed framework is robot-agnostic and thus the state-space could potentially be different.
%Thus, the state-space for a problem with $12$ movable object would be $100$-dimensional.

% Although, the domain of the state-variables is continuous and infinite, the state-space \states{} becomes finite using a sampling approach. 
% The state and the action spaces are related since the states that the planner can potentially visit are limited by the possible grounded actions given an initial state (actions connect problem states and are the only means by which the robot and objects can change internal state and location).

\subsection{Action Schemas $Act$}
Three high-level action schemas in~$Act$ are considered in $T$:

\begin{itemize}    
    \item \texttt{Pick}$(b,o,g)$ actions that, when valid, allow the robot to pick up, from a given robot location~$b$, a specific object~$o$ using a specific grasp~$g$ (see Figure~\ref{fig:grasps}), as long as the robot is not already holding any object.
    \item \texttt{Place}$(b,o,p,\sigma)$ actions that, when valid, allow the robot to place, from a given robot location~$b$, a particular object~$o$ (if it is the one being held), in a given placement~$p$ using a given SOP~$\sigma$ (Stable Object Pose, see Figure~\ref{fig:sops}).
    \item \texttt{Move-Base}$(b,b')$ actions that, when valid, allow the robot to navigate from its current position~$b$ to a different one~$b'$, dragging the held object in case there is one.
\end{itemize}
Note that actions are grounded when all the related parameters are fully specified. 
Thus, the action space comprises all the possible actions that could be obtained by the combination of the possible values of these parameters.

\subsection{Ground Actions $A(s)$ through Sampling}

Given a state~$s$, the sampling function~$A$ produces a set~$A(s)$ of sampled grounded actions by  sampling values for all the $Act$ parameters.
Note that the size of~$A(s)$ increases exponentially with the number of movable objects and polynomially with the sampled parameters.
The parameter values are randomly sampled seeking both a uniform density and maximum coverage of the parameter-space, taking into account that: 1)~the set of sampled placements $p$ must contain the poses of the movable objects at $s$ and their goal position(s) in $G(s)$; 2)~the placements~$p$ can only be on the tables; and, 3)~the robot locations~$b$ must be at distance from a table less than the radius of the arm's reach (outside this region the robot cannot manipulate any object and, besides, this useless base location increases the problem complexity).
This promotes richness of representation and the sampled $A(s)$ potentially generate (if executed) states where objects are spaced and, thus, geometric constraints are easier to satisfy.

Note that random parameter sampling is computationally advantageous as it efficiently explores diverse solution spaces, mitigating the curse of dimensionality by adapting to high-dimensional environments, avoiding local minima, and handling complex geometries more effectively than deterministic interpolation. Besides, using random sampling means that with enough time the algorithm ends up sampling parameter values that do allow the discrete problem to have a solution, i.e. the algorithm is probabilistically complete.

\subsection{Executability and Transition Functions $X$, $N$}
\label{sec:exec}

% \hector{Re-cast this text to make it in sync with function $X(a,s)$ in $T$ above. That is describe how $X(a,s)=\bot$ is tested, and how $X(a,s)=\rho$
% computed, where $\rho$ is motion plan that ``carries out'' grounde action $a$ in $s$.

% *** No need to talk here about *efficiency* and use of these various tests; lazy approch etc; explained from this text where it belows (optimization)}
% \hector{Existing text:}

In a given state~$s$, the executability function $X$ maps a grounded action~$a\in A(s)$ to a motion plan $\rho = X(a, s)$, i.e. a  trajectory in the robot joint-space that respects its capabilities and does not involve self-collisions or  collisions with the environment. The action is not feasible if $\rho=\bot$.
For these operations and checks,  the motion planning framework  MoveIt! is used \cite{moveit} to  compute the following three functions sequentially:
%%% although this work is framework-agnostic, to simultaneously map the actions to a real robot motion and validate them,
%% Hector: Confusing; removed. 
\begin{enumerate}
    \item \emph{InArmWorkspace}: Returns \texttt{true} when a given target (e.g.~an object to interact with) is within the robot arm's reach, and \texttt{false} otherwise (see Figure~\ref{fig:in_workspace}).
    
    \item \emph{InverseKinematics}: Returns \texttt{true} only when a valid non-collision robot joint configuration is found, to place the robot gripper at a target (see Figure~\ref{fig:inverse_kinematics}), using a general robot-agnostic iterative Inverse Kinematics~(IK) solver algorithm~\cite{IK}.
    Notice that if the target is not within the arm workspace, i.e.~\mbox{\emph{InArmWorkspace}} fails, an IK solution does not exist, and \emph{InverseKinematics} will fail too. 

    \item \emph{MotionPlan}: Returns \texttt{true} only when a valid motion plan~$\rho \neq \bot$ is found, to place the robot gripper in a target and perform the corresponding action (see Figure~\ref{fig:motion_plan}), using a motion-planning algorithm.
    Here, RRT-Connect is used~\cite{RRTConnect}, which is probabilistically complete (if there exists a motion plan, it will find it provided sufficient time).
    Note that if there is no valid robot configuration to grasp an object, i.e.~\emph{InverseKinematics} fails, a motion-plan solution does not exist and, hence, \emph{MotionPlan} will fail after a timeout.
\end{enumerate}

The state-transition function~$N$ returns the last state~$s'$ of the motion plan~$\rho\neq\bot$.

%% \hector{Meaning of these two sentence not clear:} Notice that the introduced validation-pipeline refers only to manipulation actions. For other cases, other validation steps may be required.
%% Hector: Not clear, removed. 

\section{Solving TAMP: Sketch Decompositions}

We will  search for plans that solve a given TAMP task  \mbox{$T = \tup{S,s_0,G,Act,A,X,N}$} by means of the
the \siwR{}  algorithm that uses a domain-dependent sketch $R$ to decompose the problem into subproblems, and
the IW search algorithm for solving the subproblems. If the width of the sketch~$R$ for the target class of problems~$\Q$ is bounded, then \siwR\ will solve the problems in polynomial time, and moreover, if the sketch width is~$1$,
\siwR\  will solve the subproblems, greedily with the IW(1) algorithm, in linear time.
For making this possible, the  features used in  the sketch~$R$, and the features used in the IW algorithm,
need to be chosen carefully, assuming that the class of problems~$\Q$ can be solved in polynomial time.
%It has been shown that sketches~$R$ bounded by a given width parameter~$k$, along with the features involved, can be both learned without supervision, for all instances~$\Q_D$ of classical planning domains~$D$, from a small number of small domain instances~\cite{drexler2022learning}.
For the TAMP setting, we provide the sketch and the features by hand, although they could be learned from small instances~\cite{drexler2022learning}, provided suitable \emph{state languages} (in planning, the state languages is given by the domain predicates).

The set of features~$F$ is given by two sets~\mbox{$F=\tup{F_H,F_D}$} of Boolean and numerical state features (functions): 
the first set~$F_H$ is used only in the IW searches invoked by the \siwR\ algorithm and they are just Boolean features; 
the second set~$F_D$ is used in the sketch~$R$ for expressing the decomposition of a TAMP problem~$T$ into subproblems,
and uses both Boolean and numerical features. The numerical features \emph{counters} take non-negative integer values.
The abstract state space that is searched is actually given by the possible values of the features. It is assumed
that the features for the class of problems~$T$ are rich enough to distinguish goal states from non-goal states.

\subsection{Features $F_H$ for Searching Subproblems}

Given a sketch with rules $C \mapsto E$, the subproblem to be solved at a state $s$ is given by a problem that is like $T$
but with initial state $s$ and goal states $G_R(s)$ given by the original goal states of the problem $T$,
along with the ``subgoal'' states $s'$ such that the state pair $[s,s']$ satisfies a sketch rule $C \mapsto E$ in $R$; namely,
$s$ makes the condition $C$ \texttt{true}, and the transition from $s$ to $s'$ makes the effect expression $E$ \texttt{true} (for instance,
a feature is decremented). These  subproblems, denoted as $T[s,G_R(s)]$, are solved by an IW(1) search
where the ``atoms''  in a state $s$ are given by the Boolean search features $F_H$. The  IW(1) search
over the subproblem $T[s,G_R(s)]$ thus proceeds as a breadth-first search from $s$ where
newly generated states that do not make a feature in $F_H$ true for the first time. The search
succeeds when a state $s'$ in $G_R(s)$ is found, else it fails. The applicability of an action $a \in A(s)$
is checked by evaluating the executability function $X(a,s)$: if it results in a motion plan $\rho$,
the state resulting from the action $a$ in $s$ is set to $s'=N(s,\rho)$. 

The set~$F_H$ of search features considered capture the robot and objects' poses
and they are called  \texttt{robot-at-$p$} and \texttt{object-$o$-at-$p$}, for
poses~$p$ and objects~$o$.
It can be shown that if the subproblems $T[s,G_R(s)]$ generated by the sketch $R$
can all be solved by moving an object from some original pose to another pose, the width of the subproblems with these features~$F_H$ is $1$, 
and thus they can be solved efficiently by IW(1), while generating a number of
nodes that is linear in the number of objects, sampled configurations,
and number of features in~$F_H$. 

\subsection{Sketch, \siwR\ and  Problem Decomposition}
\label{sec:sketch}

Starting with a state $s$, the \siwR\ algorithm carries out an IW search from $s$ until a state subgoal state  $s'$ in $G_R(s)$ is found.
If $s'$ is a goal state of the problem, the \siwR\ exits successfully, else it repeats the process from $s := s'$. 
The role of the sketch $R$ is to define the set of subgoal states $s'\in G_R(s)$ as the goal states of the problem, and the subgoal states $s'$
that along with $s$ satisfy a sketch rule.

For the different types of pick-and-place tasks to be addressed, a hand-made sketch $R$ with four rules involving five features is used. 
The five features are:

\Omit{
The algorithm $\text{SIW}_R$ is able to decompose the problem into sequential subproblems using a user-provided domain knowledge, encoded into a set of rules that outline a sketch of the solution. 
The proposed sketch captures the following strategy. Given that the goal is that no object is misplaced, whenever a misplaced object is accessible (either to pick it up or to leave it) it should be moved to a place where it is not misplaced. If there is no accessible misplaced object, an action should be done to help making any object more accessible (without disturbing the others). 
}

\begin{itemize}
    \item $H$: a Boolean feature that is  \texttt{true} if the robot is holding an object, and \texttt{false} otherwise.

    \item $I$: a Boolean feature  that is  \texttt{true} if there exists a goal pose of the grasped object not blocking the pick/place of any other misplaced object, and \texttt{false} otherwise.

    \item $m$: the number of \emph{misplaced} objects, defined as the objects that are standing (i.e.~not held) outside their goal position/region (if the object has no associated goal, it is never misplaced) or are held and their goal position/region is blocked.
    The goal positions/regions depend on the problem and they can be defined in absolute coordinates (e.g.~\emph{Sorting}) or relative coordinates (e.g.~\emph{Words}).
        
    \item $u$ and $v$: respectively $\min_{i=1,...,m}(\alpha_i\!+\!\beta_i)$ and $\sum_{i=1,...,m} \alpha_i$, with $\alpha_i$ being the minimum number of objects blocking the $i$-th misplaced object (i.e.~preventing it from being picked up from its current position and put down in its goal position/region) and $\beta_i$ the minimum number of misplaced objects blocked by the goal position(s) of the $i$-th misplaced object.
    During the search for such a minimum value, it is considered the best robot location and grasp for the \emph{pick} action and the best robot location, object placement and SOP for the \emph{place} action in the goal zone, while maintaining consistency (i.e.~using the same grasp for the considered \emph{pick} and \emph{place} pair).
\end{itemize}
Using these features, the goal in all the tasks is expressed as $\neg H$ and $m=0$ (no misplaced objects and no held object).

The sketch $R$ used is made up of four sketch rules explained below, following the notation of the work that first introduced the sketches~\cite{bonet-geffner-aaai2021}.
%Recall that the antecedent~$C$ of a sketch rule $C \mapsto E$ involve Boolean conditions while the effect expressions are either Boolean literals or numerical expressions of the form $\DEC{m}$ or $\INC{m}$ that express that the value of $m$ has be decreased or increased respectively. These effect expressions define the subgoals to be achieved. Effect expressions of the form $m?$ indicate that it . Otherwise, a non-mentioned feature must preserve its value. 
Recall that $\UNK{m}$ indicates that it does not matter how $m$ changes, and a non-mentioned feature must preserve its value.

% \begin{itemize}
%   \item $\{\neg H,\, m>0,\, u=0 \} \mapsto \{H,\, F,\, \DEC{m},\, u?,\, \neg\,\INC{v}\}$:
%   %\item $\{\neg H,\, m>0,\, u=0 \} \mapsto \{H,\, F,\, m\!\downarrow,\, u?\}$:
%       if there is some misplaced object ($m>0$) that can be picked up and placed down directly ($\neg H$ and $u=0$),  pick it up (i.e.~lower~$m$ and make~$H$ \textrm{true}).
    
%     \item $\{ \neg H,\, m>0,\, u>0 \} \mapsto \{ H,\, F?,\, m?,\, u?,\, \DEC{v} \}$:
%       if there are misplaced objects ($m>0$) but they cannot be picked up directly ($\neg H$ but $u>0$),  pick up an  obstructing object (i.e.~decrement~$v$).
%     However, even when $u$ were inaccurately estimated, picking an unexpectedly-accessible misplaced object would also satisfy this rule (which is not an inconvenience at all).

%     \item $\{ H,\, \neg F \} \mapsto \{ \neg H,\, F?,\, m?\}$:
%     if  robot is holding an object that cannot be placed in  goal without blocking misplaced objects,   place it somewhere else without creating new interferences with the objects that are still  misplaced  ($m$ may increase but  $u$ and  $v$ should not change).
    
%     \item $\{ H,\, F \} \mapsto \{ \neg H,\, F? \}$:
%     if the robot is holding an object, which would not block other misplaced object once placed if it had a goal associated with it, enforcing to put it down but not in a wrong position (i.e.~do not change~$m$) and without disturbing the access to the pending misplaced objects (do not change $u$ or $v$).
% \end{itemize}
\begin{itemize}
  \item $\{\neg H,\, m>0,\, u=0 \} \mapsto \{H,\, F,\, \DEC{m},\, u?,\, \neg\,\INC{v}\}$:
  %\item $\{\neg H,\, m>0,\, u=0 \} \mapsto \{H,\, I,\, m\!\downarrow,\, u?\}$:
    if there is some misplaced object that can be picked up and placed down directly,  pick it up.
    The expression  $\neg\,\INC{v}$ means that  $v$ can change but not increase. 
    
    \item $\{ \neg H,\, m>0,\, u>0 \} \mapsto \{ H,\, I?,\, m?,\, u?,\, \DEC{v} \}$:
      if there are misplaced objects but they cannot be picked up directly,  pick up an  obstructing object.
      Notice that, even when~$u$ were inaccurately estimated, picking an unexpectedly-accessible misplaced object would also satisfy this rule (which is not an inconvenience at all).

    \item $\{ H,\, \neg I \} \mapsto \{ \neg H,\, I?,\, m?\}$:
    if  robot is holding an object that cannot be placed in  goal without blocking misplaced objects,   place it somewhere else without creating new interferences with the objects that are still  misplaced.
    
    \item $\{ H,\, I \} \mapsto \{ \neg H,\, I? \}$:
    if the robot is holding an object, which would not block an other misplaced object once placed if it had a goal associated with it, enforcing to put it down but not in a wrong position and without disturbing the access to the pending misplaced objects.
\end{itemize}
The sketch structures the problem into subproblems but it is not a policy; the subproblems need to be solved by search, but this is a search
that  is done efficiently by  IW(1)  in \siwR.  For example, the rules do not say how the base of the robot should move, or which grasp to use.
Part of these details are filled in by the search (e.g.~robot base moves) and the others by the motion planner (e.g.~grasps).
Besides, note that while the approach is general, the given sketch is suited for (a large class of) pick-and-place tasks. For TAMP tasks involving other kinds of motions such as drag, push, etc., the sketch can be reused, but new features would need to be implemented accordingly.

%\subsubsection{Lazy action-validation}

\medskip

\noindent\textbf{Formal Properties.}
The effectiveness of the {\siwR} procedure on the TAMP tasks~$T$ considered below depends on the problem decomposition that follows the choice of the sketch~$R$ (and its features~$F_D$), and the  power of the subproblem search by means of IW(1) with the given search features~$F_H$. Two properties that  would guarantee that {\siwR} using  IW(1) solves the problem $T$ are: 1)~sketch termination~\cite{bonet-geffner-aaai2021}, and 2)~sketch width of $1$ for this family of tasks.
We prove the first and provide a rationale for the second, that in the TAMP setting involves conditions that cannot be captured with precision.

\medskip

\noindent \textbf{Sketch termination.}
It must ensure that moves from state~$s$ to subgoal state $s'\!\in\!G_R(s)$ cannot go on for ever.
For this, we show that the four sketch rules can be used to generate subgoals a finite number of times only.
First, note that feature~$v$, that captures roughly total number of ``interferences'' towards the goal, is decreased in the second sketch rule~$r_2$ but is not increased in any of the four rules (no expression effects $\INC{v}$ or $\UNK{v}$).
This means that rule~$r_2$ can be used for subgoaling a finite number of times only  as  counters cannot become negative.
Once that rule~$r_2$ is excluded from those that can be ``followed'' an infinite number of times, a similar argument
can be made about rule~$r_1$ where the counter~$m$ is decreased.
In this case, rule~$r_3$ may increase~$m$ as it contains the effect expression~$m?$, yet this rule cannot be followed after $r_1$ because $r_1$ makes~$I$ \texttt{true} and $r_3$ requires $I$ \texttt{false}.
So infinite ``executions'' of rule~$r_1$ can only involve infinite executions of rule~$r_4$, as rule $r_2$ cannot be executed infinitely often, and since $r_4$ cannot increase the value of~$m$, $r_1$ cannot be executed infinitely often, but then neither $r_3$ nor $r_4$ that make $H$ \texttt{false} but require $H$ \texttt{true} in the antecedent.

\medskip

\noindent \textbf{Sketch width 1.} The width of the presented sketch for the class of pick-and-place problems $T$
with search features~$F_H$ can be shown to have width~$1$ under the following conditions: 1)~the counters  $m$,  $u$, and $v$ represent all interferences affecting misplaced objects; 2)~there is reachable space for placing objects without causing new interferences (increases in $u$ and $v$);
3)~every object can be reached, possibly after moving other objects out of the way,
4)~the ground parameters (sampling) of the relevant ground actions in the subproblem  are sampled. 
These are strong assumptions but  they are needed for explaining the power and scope of the proposed method.
It is important that the sampling is done in the context of a simple subproblem, and does not need to be done once for the whole problem.
We show  then that in every state $s$, the problem
of reaching a subgoal state $s'$ $G_R(s)$ has width $1$.
The four sketch rules $r_i: C_i \mapsto E_i$ have disjoint antecedents $C_i$,
so we just need to consider four cases, where $s$ makes $C_i$ \texttt{true}, and $s' \in G_{r_i}(s)$ where the set of rules $R$ is replaced by the single rule $r_i$, $i=1, \ldots, 4$.

\noindent \emph{Rule~1.} If $s$ satisfies $C_1$,  the subproblem $T[s,G_{r_1}(s)]$ involves picking up a misplaced object with no obstructions,
whose goal is free and can be occupied without causing obstructions. The rule ensures that there is one such object and that picking up an object that does not comply  with these conditions will not lead to a state $s'$ in $G_{r1}(s)$. Therefore, there is a plan to solve this subproblem in which the robot moves certain times and the object is picked up.
  If this plan is optimal, the sequence of robot base positions, until the pick up, captured by \texttt{robot-at-$p$} features,  followed by the object configuration after the pick up, captured by the  \texttt{object-$o$-at-$p$} features, yields chain of ``atoms'' $t_0, \ldots, t_k$ that is \emph{admissible};
  namely, $t_0$ is \texttt{true} in $s$, optimal plans for $t_i$ can all be extended into optimal plans for $t_{i+1}$, and optimal plans for $t_k$ are optimal solutions to the subproblem $T[s,G_{r_1}(s)]$. Hence, IW(1) running with the proposed $F_H$ will solve this subproblem optimally~\cite{lipovetzky-geffner-ecai2012,bonet-geffner-aaai2021}.
  
\noindent   \emph{Rule~2.}  If $s$ satisfies $C_2$,  the subproblem $T[s,G_{r_2}(s)]$ involves picking up an object that is obstructing misplaced objects, for moving it  ``out of the way''. The subproblem can be solved in the same way by \texttt{Move-Base} actions,
  followed by a \texttt{Pick}, giving rise to a similar admissible chain of atoms that establishes that the subproblem has width~$1$ and is solved optimally by IW(1).
  
\noindent   \emph{Rule~3.}  If $s$ satisfies $C_3$,  the subproblem $T[s,G_{r_3}(s)]$  involves placing the object held ``out of the way'', without increasing the number of interferences. The optimal plan for the subproblem may involve a number of robot moves followed by placing down the held object. In this case,
  the ``atoms''  \texttt{object-$o$-at-$p$} that are made \texttt{true} in an optimal plan for the subproblem constitute an admissible chain
  that proves that the subproblem has width~$1$.
  
\noindent   \emph{Rule~4.} If $s$ satisfies $C_4$,  the subproblem $T[s,G_{r_4}(s)]$ involves placing the object held, that appears in the goal, whose target is not obstructed and that placed in that target will not cause obstructions to misplaced objects. Once again, the conditions ensure that the subproblem has a solution, and potentially involves \texttt{Move-Base} actions followed by a \texttt{Place} action as well.
  The ``atoms''  \texttt{object-$o$-at-$p$} that are made \texttt{true} in an optimal plan for the subproblem constitute an admissible chain, that proves
  that the subproblem has width~$1$ (under the above, general conditions).

\subsection{Optimizations}

Here, we introduce implementation details, including used approximations and optimizations.
There are two sources of approximation: 1)~Representation of continuous variables through sampling and, 2)~Relaxing the Sketch-Features computation.
Furthermore, two search optimization mechanisms are introduced that focus on minimizing the efforts on geometric validation of the actions.
 
\medskip

%\subsubsection{Adaptative sampling and Probabilistic completeness}
\noindent\textbf{Adaptive sampling and Probabilistic completeness.}
Since the sketches allow decomposing the problem into subproblems, we use a different state-space adapted to each subproblem.
This enables working with reduced tractable state-spaces that represent well each subproblem scenario.
Note that any of these specific discretizations is not sufficiently-rich to represent well the whole problem world and yet, combined, they allow finding a solution to the overall problem.
Besides, the sampling density of the $Act$ parameters is increased (and the search is restarted) when an attempt to reach a subgoal fails, to obtain new values that can help to find a subproblem solution.
Thus, the full approach is probabilistically complete in the sense that (a)~all object locations and configurations are susceptible to be sampled; (b)~when a subsearch fails, the same subsearch is reattempted but with more sampled object locations.

\medskip

%\subsubsection{Sketch-Features computation}
\noindent\textbf{Sketch-Features computation.}
The computation of the features $I$, $u$ and $v$ would imply complex inverse-kinematics and collision-checking computations.
To speed up the process, these type of computations are avoided and approximations are used instead in the approach implementation.
In particular, instead of finding the robot configuration that causes the fewest collisions for a given pick/place of a misplaced object, we overestimate this number by looking at how many objects have their center of mass within a 3D region containing all possible inverse kinematics solutions. 
This region is easy to compute and is implicitly defined considering the object and robot poses and the grasp to be used. 

\medskip

%\subsubsection{Lazy action-validation}
\noindent\textbf{Lazy action-validation.}
To speed-up the search, the action validation is relaxed until a subplan is found.
In particular, the third step of the executability function~$X$ (i.e.~\emph{MotionPlan}) is only checked for the actions in a potential subplan.
Hence, graph edges are ``provisional'' until complete checking and we must keep all the discovered edges.
Thus, a node may have more than one parent node (although the one implying a lowest cost to reach the start is the one acting as the parent).
With this lazy approach, the most time consuming action-validation step is performed only on those actions that are part of a potential solution plan.
If all the plan actions are successfully validated, a completely valid subplan is returned. 
Else, the unfeasible action is discarded and  the corresponding graph edge is removed.
However, the child node is kept as long as there is an edge supporting it.

\medskip

%\subsubsection{Incremental \iw{k}}
\noindent\textbf{Incremental \iw{k}.}
Instead of restarting completely the \iw{k} subsearch when 
a potential subplan is invalidated by the motion planner, we save the search done until then and
resume the search.
For this, the novelty management is adapted such as follows:
1)~All the features associated to each generated node are recorded (it could still be pruned, if it does not introduce the required novelty);
2)~A feature is supported by one main node but it may have newer nodes as backup candidates;
3)~When a node has to be discarded because it has been disconnected from the root node during a final action-validation, this node is removed as support of its related features;
4)~If a feature loses all of its supporters it is removed from the novelty table; and,
5)~A pruned node (because of not passing the novelty check) could be recovered if it is the next backup candidate after removing a support node. 

The resulting algorithm after applying the explained optimization is named in this work Lazy Serialized Incremental Iterated Width with Sketches (\textbf{Lazy-SIIW$_R$}).

%% file: sections/related_work.tex
\section{Related Work}\label{sec:rel}

Our approach to TAMP interleaves  task and motion  plans like \cite{doi:10.1146/annurev-control-091420-084139}%,cambon_hybrid_2009
. 
Two other related  approaches are  \emph{sequence-before-satisfy}, that  first obtain fully-symbolic potential plans and then solve
the geometric constraints~\cite{garrett_pddlstream_2020}%,Shoukry_SMC,bidot_geometric_2017
, and
\emph{satisfy-before-sequence}~\cite{7733599}, that  first solve the geometric problem
% to find sets of satisfying assignments for individual constraints
and then find  action sequences that use those values.

%Other possible options are obtaining fully-symbolic potential plans and then solve the geometric constrains, i.e.~\emph{sequence-before-satisfy} approaches~\cite{garrett_pddlstream_2020,Shoukry_SMC,bidot_geometric_2017}, or solving first the geometric problem to find sets of satisfying assignments for individual constraints and, afterwards, try to find action sequences that use those values, i.e.~\emph{satisfy-before-sequence}~\cite{7733599}.
Our work is also related to Planet~\cite{Thomason}, which interleaves task and motion planning through a flexible sampling-based approach. Similarly, Planet avoids sampling in the full Cartesian space.
However, Planet uses a composite space including symbolic and continuous components, and defines an explicit embedding of the symbolic state into the continuous space.
Furthermore, Planet uses a heuristic to guide the sampling. Instead, we consider a  blind but focused search, IW(1), to solve each subproblem,
and the subproblem to solve next follows from the sketch.
% uses the same benchmark as we do in this paper~\cite{Lagriffoul2018PlatformIndependentBF}.
An interleaved approach using sampling mechanism of predefined primitives to obtain (only valid) continuous values in the node expansion is proposed in~\cite{AJANOVIC2023105893}, where a feasibility map is proposed to enable approximated models for motion primitives generation.

\Omit{
A majority of TAMP approaches use PDDL to encode the problem as in pure task-planning problems.
Modeling the problem with PDDL can be done with the basic formulation of this language. 
However, there some works that use extended versions of this language. 
In \cite{FerrerMestres2017CombinedTA}, PDDL is extended using Functional
STRIPS \cite{Geffner2000_functional_strips}, that is able to accommodate procedures and state constraints.
State constraints represent implicit action preconditions to discard spatial overlaps.
Procedures are used for testing and updating robot and object configurations.
Furthermore, in \cite{garrett_pddlstream_2020}, it is introduced a PDDL extension that introduces streams as an interface for incorporating sampling procedures as a black box.
It enables conditionally sampling values (depending on existing values) such as robot configurations that satisfy geometric constraints.
}

%Most of TAMP approaches use extensions of PDDL for encoding the problem.
Most  related to our approach is the work of \citeauthor{FerrerMestres2017CombinedTA}~(2017) which uses Best-First Width-Search (BFWS) on a pre-discretized state-space and restricts the sampling to a template-grid (kept for the whole search). BFWS  uses an extension of PDDL that accommodates procedures and state constraints \cite{Geffner2000_functional_strips}.
State constraints represent implicit action preconditions to discard spatial overlaps.
Procedures are used for testing and updating robot and object configurations.
Other works extend PDDL in different ways. For example, PDDLStream~\cite{garrett_pddlstream_2020} incorporates sampling procedures in PDDL that allow a planner to reason about conditions on the inputs and outputs of a conditional generator while treating its implementation as a black box.

Using sketches to address a collection of tasks have parallelisms with generalized planning~\cite{Srivastava2008LearningGP}%,Bonet2009AutomaticDO,10.5555/3038594.3038630
.
Like general policies~\cite{bonet-geffner-ijcai2018}, sketches are general and not tailored to specific instances of a domain, but unlike policies, the feature changes expressed by sketch rules represent sub-goals that do not need to be achieved in a single step.
A methodology for learning sketches in classical planning tasks has been introduced in~\cite{drexler2022learning}. Learning features, abstractions, and generalized plans from a few examples in continuous TAMP problems has been proposed by~\citeauthor{curtis2022discovering} (2022).

%streams as an interface for incorporating sampling procedures as a black box.
%It enables sampling of robot configurations that satisfy geometric constraints conditionally on  existing values) .

%Also check
%Dmitry Berenson, Siddhartha S. Srinivasa, and James Kuffner. “Task Space Regions: A
%Framework for Pose-Constrained Manipulation Planning”. In: International Journal of
%Robotics Research 30.12 (2011), pp. 1435–1460.

%Caelan Reed Garrett, Tomás Lozano-Pérez, and Leslie Pack Kaelbling. “Sampling-Based
%Methods for Factored Task and Motion Planning”. In: The International Journal of Robotics
%Research 37.13-14 (2018), pp. 1796–1825.

%In this work we do not use PDDL instead, avoiding explicitly formulate the domain and interestingly not declaring explicitly the action structure (and working directly with a simulator). Other works avoid using PDDL by performing normal motion planning in a space including both geometric and symbolic states \emph{Planet}~\cite{Thomason}.

%We use an adaptive and continuous sampling while, for instance, \emph{BFWS}~\cite{FerrerMestres2017CombinedTA}, which also use a width-based algorithm,
%restricts the sampling to a template-grid (kept for the whole search). 

The idea of lazy evaluation is known to reduce planning time for search-based planners and has been applied recently in several works, e.g,
by postponing applicability checks for successor states performing geometric queries~\cite{dornhege_lazy_nodate}, by
deferring motion sampling until an action skeleton is found \cite{khodeir2022policy}, or
by solving shortest path problems as needed \cite{dellin_unifying_2016}.

An alternative formulation leverages nonlinear optimization to jointly compute a motion that satisfies geometric and physical constraints~\cite{Ortiz_Karpas_Toussaint_Katz_2022}%,10.5555/2832415.2832517,Braun2021RHHLGPRH
. Contrary to our approach, these methods operate on the full planning sequence and do not explicitly exploit the subproblem decomposition. Direct comparison with our approach is difficult, as they require the formulation of the optimization problem, e.g.~the logic-geometric program, and a full PDDL problem-description. 

Hierarchical planning approaches also express and exploit problem decompositions \cite{wolfe2010combined,kaelbling2011hierarchical}.
In this cases, the decomposition is expressed in hierarchical manner and not by means of sketches.
The key difference between these hierarchical approaches and the sketch-based approach
is the role and scope of the search. In the proposed sketch-based approach, the subproblem
search is done in linear time, by means of the IW(1) algorithm, and this guaranteed
to be complete if the sketch width is 1. The role and scope of the search in hierarchical
approaches in both robotics and planning is less clear: either the domain knowledge must
express a full strategy for solving the problems which does not require any search,
or else, the search can easily get lost. More recent works make use of hierarchical 
models and combine them with learning \cite{patra2020integrating}.

\Omit{
  %% Hector: the description is not sufficiently clear; removed and replaced by above text. When not clear, at least be concise.
Hierarchical planning approaches also exploit the subproblem decomposition. 
The hierarchization of the problem can be specified by hand \cite{kaelbling2011hierarchical}, and then take decisions and commit to them in a top-down fashion. 
This limits the length of the plans that need to be explored and thus the search effort is dramatically decreased. 
Another related approach makes use of Hierarchical Task Networks~\cite{wolfe2010combined}. For this, an action-hierarchical structure must be defined in which, during the planning, abstract actions are refined into executable actions.
More recent works make use of hierarchical operational models to perform tasks in changing environments searching in the operational space of those models~\cite{patra2020integrating}. 
Furthermore, they apply a learning strategy to acquire, from experience, a mapping from decision contexts to method instances and an heuristic to guide the search.
}

Other related methods  make  flexible use of external predicates and  functions for feasibility checks~\cite{erdem15,Esra:11}. These methods strongly rely on the problem description, which is modified during the search integrating additional domain specific information from
the motion-level. %  when a motion planning failure occurs due to some infeasibility not captured by geometric reasoning handled by external predicates/functions.
\Omit{
Width-based does not steer the search towards the goal but others use heuristics~\cite{Esra:11}.}
Learning-driven approaches can be used to guide the solution of TAMP problems. 
\cite{pmlr-v100-kim20a} proposed learning action-value functions for speeding up the discrete part of the search in TAMP problem.
Another learning approach, introduced in \cite{mcdonald_guided_2022}, consists on training a policy to imitate a TAMP solver’s output.
The obtained feed-forward policy is used to solve tasks from sensory data and to supervise the training an asynchronous distributed TAMP solver for imitation learning is used.

\Omit{  not a good parargraph, no space either
  
Finally we assume deterministic actions but planning approach could be extended to solve POMDPs leveraging learning approaches~\cite{Zhang:13}. 
In this work, the authors make a POMDP planning-problem tractable using a hierarchical decomposition of the POMDP.
%Interestingly, making tractable complex TAMP problems trough decomposition is a strategy shared with our work.
In a way, they also make tractable complex TAMP problems trough decomposition.
Note however, that our framework plans fully online and does not assume any predefined template for the positions of objects, nor predefined object-characteristics (e.g. object shape), thus, via replanning, it could handle sensory and action uncertainty. 
}

%minimal infeasible subgraph-NLP
%These conflicts are found during the attempts to solve the entire problem 
%attempting to solve the global problem  to identify and refine the PPDL description of the problem.
%reformulates PDDL. While, we do not require a PDDL actions.
%Our approach subproblems sampling 

%% file: sections/experimental_results.tex
\section{Experimental Results}\label{sec:results}
% \url{https://drive.google.com/drive/folders/110VtyoSNrw0hjR7Wma7fumqJIB9wxr3L}
We evaluate our method in the three tasks\footnote{Supplementary material includes accompanying videos.} 
presented initially
% in Section~\ref{sec:tasks}
% Hector: Removed: label doesn't seem to work 
with the following objectives in mind: to provide empirical evidence of the theoretical results, to analyze how the method scales with the problem complexity, to quantify the impact of the lazy action-validation, and to compare it with related approaches. We compare the following methods:
\begin{itemize}
\item \textbf{Lazy-SIIW$_R$}: our approach with lazy action-evaluation, called Lazy Serialized Incremental IW with Sketches.
\item \textbf{SIW$_R$}: our approach without lazy action-evaluation.
\item \textbf{BFWS}~\cite{FerrerMestres2017CombinedTA}. 
\item \textbf{Planet}~\cite{Thomason}.
\item \textbf{PDDLStream}~\cite{garrett_pddlstream_2020}. We extended the existing implementation of PDDLStream planners. Details of these extensions are provided in the supplementary material.  
\end{itemize}

\begin{table*}[t!]
    \newcommand{\STAB}[1]{\begin{tabular}{@{}c@{}}#1\end{tabular}}
    \caption{
        Average results over ten runs in the benchmarking problems in an Intel{\small{\textregistered}}~Core\texttrademark~\mbox{i7-10610U CPU} at 1.80~GHz, with 16~Gb of RAM, on Ubuntu 20.04.4 and ROS Noetic.
        The results of BFWS and Planet are respectively taken directly from \citeauthor{FerrerMestres2017CombinedTA} (2017) and \citeauthor{Thomason} (2022).
        $^\dagger$Within $30~\si{min}$ maximum planning time or maximum memory exceeded.
        $^\perp$Considering that the path execution does not start until the path has been completely planned.
        $^*$Including also the pre-processing time, for a fair comparison.
      }\label{tab:solution}
      \small
    \centering
    \setlength\tabcolsep{1mm}
    \begin{tabular}{|c|c|c|c|c||c|c|c|c|c|c|c|}
    \hline
    
    \makecell{Prob.} &
    \#Tables &
    \#Objects &
    \makecell{\#Goal\\objects} &
    \makecell{Clutter\\level} &
    Planner &
    \makecell{Success\\ratio$^\dagger$} &
    \makecell{Planning\\time} &
    \makecell{Execution\\time} &
    \makecell{Total\\time$^\perp$} &
    \makecell{\#Expanded\\nodes} &
    \makecell{\#Sub\\plans} \\ \hline\hline
    
    \multirow{20}{*}{\STAB{\rotatebox[origin=c]{90}{Sorting Objects}}} &%    \multirow{9}{*}{\STAB{\rotatebox[origin=c]{90}{Sorting Objects}}} &
    \multirow{3}{*}{1} &
    \multirow{3}{*}{20} &
    \multirow{3}{*}{2} &
    \multirow{3}{*}{High} &
    \textbf{Lazy-SIIW}$_{\boldsymbol{R}}$ &
    100\% &
    3.16~\si{\minute} &
    3.67~\si{\minute} &
    6.83~\si{\minute} &
    60 &
    14 \\ %\cdashline{6-12}

    &
    &
    &
    &
    &
    SIW$_R$ &
    100\% &
    13.05~\si{\minute} &
    3.73~\si{\minute} &
    16.78~\si{\minute} &
    54 &
    14 \\ %\cdashline{6-12}

    &
    &
    &
    &
    &
    BFWS &
    100\% &
    5.25~\si{\minute}$^*$ &
    5.91~\si{\minute} &
    11.16~\si{\minute}&
    63.3k &
    1 \\ \cline{2-12}

    &
    \multirow{9}{*}{3} &
    \multirow{5}{*}{2} &
    \multirow{5}{*}{2} &
    \multirow{5}{*}{Low} &
    \textbf{Lazy-SIIW}$_{\boldsymbol{R}}$ &
    100\% &
    0.26~\si{\minute} &
    1.21~\si{\minute} &
    1.47~\si{\minute} &
    28 &
    4 \\ %\cdashline{6-12}

    &
    &
    &
    &
    &
    PDDLStream \emph{Adaptive} &
    100\% &
    0.03~\si{\minute} &
    1.24~\si{\minute} &
    1.27~\si{\minute} &
    N/A &
    1 \\ %\cdashline{6-12}

    &
    &
    &
    &
    &
    PDDLStream \emph{Binding} &
    100\% &
    0.21~\si{\minute} &
    1.20~\si{\minute} &
    1.41~\si{\minute} &
    N/A &
    1 \\ %\cdashline{6-12}

    &
    &
    &
    &
    &
    PDDLStream \emph{Incremental} &
    100\% &
    12.01~\si{\minute} &
    1.22~\si{\minute} &
    13.23~\si{\minute} &
    N/A &
    1 \\%\cdashline{6-12}
&
    &
    &
    &
    &
    PDDLStream \emph{Focused} &
    0\% &
    N/A &
    N/A &
    N/A &
    N/A &
    N/A \\ 

     \cline{3-12}

    &
    &
    \multirow{4}{*}{25} &
    \multirow{4}{*}{5} &
    \multirow{4}{*}{Med.} &
    \textbf{Lazy-SIIW}$_{\boldsymbol{R}}$ &
    100\% &
    3.36~\si{\minute} &
    3.95~\si{\minute} &
    7.31~\si{\minute} &
    235 &
    10 \\ %\cdashline{6-12}

    &
    &
    &
    &
    &
    SIW$_R$ &
    100\% &
    21.60~\si{\minute} &
    4.02~\si{\minute} &
    25.62~\si{\minute} &
    198 &
    10 \\ %\cdashline{6-12}

    &
    &
    &
    &
    &
    BFWS &
    100\% &
    13.47~\si{\minute}$^*$ &
    7.61~\si{\minute} &
    21.08~\si{\minute} &
    3.5k &
    1 \\ %\cdashline{6-12}
    
    &
    &
    &
    &
    &
    PDDLStream (Any) &
    0\% &
    N/A &
    N/A &
    N/A &
    N/A &
    N/A \\ \cline{2-12}

    &
    \multirow{7}{*}{4} &
    \multirow{4}{*}{7} &
    \multirow{4}{*}{7} &
    \multirow{4}{*}{Low} &
    \textbf{Lazy-SIIW}$_{\boldsymbol{R}}$ &
    100\% &
    2.35~\si{\minute} &
    4.45~\si{\minute} &
    6.80~\si{\minute} &
    185 &
    14 \\%\cdashline{6-12}

    &
    &
    &
    &
    &
    Planet &
    100\% &
    3.22~\si{\minute} &
    N/A &
    N/A &
    N/A &
    1 \\ %\cdashline{6-12}

    &
    &
    &
    &
    &
    PDDLStream \emph{Adaptive} &
    100\% &
    7.47~\si{\minute} &
    5,72~\si{\minute} &
    13.19~\si{\minute} &
    N/A &
    1 \\ 

    &
    &
    &
    &
    &
    PDDLStream (Others) &
    0\% &
    N/A &
    N/A &
    N/A &
    N/A &
    N/A \\  \cline{3-12}

    &
    &
    \multirow{3}{*}{28} &
    \multirow{3}{*}{14} &
    \multirow{3}{*}{Med.} &
    \textbf{Lazy-SIIW}$_{\boldsymbol{R}}$ &
    100\% &
    6.52~\si{\minute} &
    9.24~\si{\minute} &
    15.76~\si{\minute} &
    630 &
    34 \\ %\cdashline{6-12}

    % &
    % &
    % &
    % &
    % &
    % SIW$_R$ &
    % 100\% &
    % 47.38~\si{\minute} &
    % 9.17~\si{\minute} &
    % 56.55~\si{\minute} &
    % 587 &
    % 34 \\ %\cdashline{6-12} 

    &
    &
    &
    &
    &
    Planet &
    0\% &
    N/A &
    N/A &
    N/A &
    N/A &
    N/A \\ 

    &
    &
    &
    &
    &
    PDDLStream (Any) &
    0\% &
    N/A &
    N/A &
    N/A &
    N/A &
    N/A\\ 

    \hline\hline

    \multirow{2}{*}{\!\makecell{\!Non-\\Mono.\!}} &
    \multirow{2}{*}{2} &
    \multirow{2}{*}{10} &
    \multirow{2}{*}{10} &
    \multirow{2}{*}{Low} &
    \textbf{Lazy-SIIW}$_{\boldsymbol{R}}$ &
    100\% &
    3.53~\si{\minute} &
    8.10~\si{\minute} &
    11.63~\si{\minute} &
    565 &
    30 \\ %\cdashline{6-12}

    &
    &
    &
    &
    &
    SIW$_R$ &
    100\% &
    17.69~\si{\minute} &
    7.96~\si{\minute} &
    25.65~\si{\minute} &
    534 &
    30 \\ \hline\hline

    %\multirow{2}{*}{\STAB{\rotatebox[origin=c]{90}{Words}}} &
    \multirow{2}{*}{\!Words\!} &
    \multirow{2}{*}{1} &
    \multirow{2}{*}{11} &
    \multirow{2}{*}{4-5} &
    \multirow{2}{*}{Low} &
    \textbf{Lazy-SIIW}$_{\boldsymbol{R}}$ &
    100\% &
    2.26~\si{\minute} &
    3.49~\si{\minute} &
    5.75~\si{\minute} &
    243 &
    13 \\ %\cdashline{6-12}

    &
    &
    &
    &
    &
    SIW$_R$ &
    100\% &
    8.07~\si{\minute} &
    3.58~\si{\minute} &
    11.65~\si{\minute} &
    216 &
    13 \\ \hline
    \end{tabular}
\end{table*}

%We consider our approach with lazy action-evaluation Lazy Serialized Incremental Iterated Width with Sketches (Lazy-SIIW$_R$), the original SIW$_R$, but also BFWS and Planet~\cite{Thomason}. Additionally, we also extended the PDDLStream planners (i.e. \emph{Adaptive}, \emph{Binding}, \emph{Focused} and \emph{Incremental}) to .

%The framework has been run in an Intel{\small{\textregistered}}~Core\texttrademark~\mbox{i7-10610U CPU} at 1.80~GHz, with 8~processors and 16~Gb of RAM, on Ubuntu 20.04.4 and ROS Noetic, and validated with a TIAGo robot from PAL Robotics (see Figure~\ref{fig:problems}).

Table~\ref{tab:solution} shows the results.
Both Lazy-SIIW$_R$ and SIW$_R$ are able to solve all problem instances.
This confirms that the assumptions regarding sketch termination and sketch width are satisfied for these benchmarks, and shows that the method remains complete even in the presence of approximations. %in the computation of the features.
Unlike all other methods, Lazy-SIIW$_R$ requires less time for \emph{computing} the plans than for \emph{executing} them, which shows the benefits of the lazy action-evaluation.% All the other methods spend more time on the former

Lazy-SIIW$_R$ and SIW$_R$ also scale up well, despite the exponential increase in the state-space due to an increase in the number of objects. 
 %Lazy-SIIW$_R$ and SIW$_R$ also scale up well with the problem complexity: increasing the number of objects, which results in an exponential increase of the state-space, does not have a dramatic effect.
%In this case, the state-space is $>1e45$ and the proposed approach allows finding a solution in linear time requiring to expand $~630$ states only. 
Compared to BFWS, Lazy-SIIW$_{R}$ is always more efficient both in planning and execution time.
Without the lazy evaluation of constraints, SIW$_R$ only performs comparably to BFWS.
PDDLStream planners and Planet are only comparable to Lazy-SIIW$_{R}$ for simple problems, but they exceed the time limit as soon as the complexity of the problems increases.
Note that we focused on showing scalability for the sorting objects task, which is the one for which comparison with alternative methods was not too difficult. As shown theoretically, we expect linear scalability for all the solved problem families (i.e.~\emph{Sorting Objects}, \emph{Non-monotonic} and \emph{Words}), since the sketch has width 1 an it decomposes the problem into subproblems that can be solved greedily in linear time.

\begin{table}[!t]
    \centering
    \caption{Profiling of the action-validation pipeline.}\label{tab:profiling}
 \small
    \setlength\tabcolsep{0.9mm}
    \begin{tabular}{|l||c|c|c|c|}
    \hline
    \multicolumn{1}{|c||}{Step / Function} & \makecell{Success} & \makecell{Timeout} & \makecell{Rel. succ.} & \makecell{Cum. succ.} \\ \hline\hline
    1. \emph{InArmWorkspace} & 35~\si{\micro \second} & N/A & 5-30\% & 5-30\% \\ 
    2. \emph{InverseKinematics\!} & 9~\si{\milli \second} & 0.15~\si{\second} & 20-40\% & 1-10\% \\ 
    3. \emph{MotionPlan} & 0.38~\si{\second} & 5~\si{\second} & 80-100\% & 0.5-10\% \\ \hline
    \end{tabular}
\end{table}

To understand better how the lazy action-evaluation affects the performance, we analyze how the computation is distributed between the different steps involved in the geometric validation of each action. 
Table~\ref{tab:profiling} shows illustrative profiling values for the three steps. Note that they act in increasing order of computational load and decreasing order of relative discriminative power (i.e.,~rejection ratio over the non-filtered actions in the previous step). This permits to efficiently discard most of unfeasible actions. Decomposing the action-validation enables delaying the computation of the most expensive step (\emph{MotionPlan}). Thus, the complete validation is only computed on those actions that are part of a potential plan. Note that satisfying the first two steps almost ensures that the action is valid.

Another important metric is the quality of the produced plans.
Although optimality can not be guaranteed, the proposed algorithm performs well in  avoiding unhelpful actions (i.e.~obtaining near-optimal solutions). 
This follows from the rules in the sketch that  discourage  moving objects that are not misplaced and do not block misplaced objects.
Hence, after moving these objects, often other goal objects became accessible without needing to move non-goal objects. 
Indeed, when comparing the results with the other approaches, plans containing fewer actions are obtained for the same problems with the proposed approach. 

High-cluttered problems are the most challenging.
On these problems, the goal objects can be obstructed by many objects, which must be moved away,
and there is a limited space where the obstructing objects can  be set aside without blocking misplaced objects. 
In this scenario, most of the inverse-kinematic~(IK) computations, which are an iterative process, reach the IK user-provided timeout without finding a solution.
Thus, the action is marked as unfeasible (even when it is valid, simply because it has not had time to find a motion that implements it).
% \hector{This is important but confusing: What do you mean that there are timeout? Does it mean that problems are not solved?
%   Not clear how timeouts imply just ``slower node expansation''. If you mean that the IW(1) search must produce more plans because
%   most plans found are found to be infeasible, say so. In any case, make following sentences crystal clear; they are not}.
This implies a slower node expansion (it is more probable to spend the whole IK time-budget in an expansion since there is not an actual solution) and a higher difficulty to find valid actions (i.e. longer total planning time).
Note that the selection of the IK time-budget is not trivial: If its extremely low, IW(1) could discard all the possible grounded actions that can lead to the goal (risking not being able to find a solution plan even when one exists) and, on the contrary, being too high implies non-affordable computational times. 
In this approach, 5~seconds has experimentally been found to keep a trade-off between finding solutions even in cluttered environments maintaining a reasonable computational-time.
In addition, as there are so few gaps, the proposed approach samples the placements more densely to ensure that there exists an accessible placement.
On the contrary, PDDLStream planners and Planet do not scale well with the number of objects. This fact is mainly explained by the fact that these approaches do not serialize the problem into smaller subproblems which allow resuming failed searches in a specific subproblem and to tailor the automatic sampling of the continuous world to the specific subproblem. On the other hand, the BFWS search  also  scales well but at the cost  of a  rigid  precompilation that  forces the objects to be  in a fixed set of  grid configurations only  during the search.
This precompilation is expensive and does not scale up that well,  but when it does so, the planner does not need to invoke the motion planner at plan time.

%% file: sections/conclusion.tex
\section{Conclusions}
\label{sec:conclusions}

We have presented an approach for TAMP that makes  use of the notions of width and sketches developed in classical planning
for decomposing  problems into subproblems. For this, a general sketch
of width $1$ has been crafted that ensures that  the families of
pick-and-place problems considered are decomposed into subproblems that can be solved greedily in linear time,
under suitable assumptions. The same sketch has been used for the  three types of tasks considered; the only
change being in the definition and computation of one of the features (number of misplaced objects).
The value of sketches is that they  allow  to specify the subgoal structure of the problems at a high-level,
without having to specify unnecessary details  that are handled by the polynomial, and in our case, linear search. 
The language of sketches is very flexible and offers  two clear  benefits in the context of TAMP.
First, when a task plan is found to be unfeasible due to the  geometric constraints, the combinatorial search
resumes in the specific subproblem where unfeasible subplan was found. 
Second, the sampling of object poses  is not done  once, globally, at the start of the search,
but locally, at the start   of each  subproblem. 
%Finally, the sketches provide  a convenient interface for integrating the functionality of the task and motion planners  as they split  the problem into subproblems, some of which are to be solved by the former and some by the latter. 
The bounded  width of the sketch provides a bound on the complexity of the subproblems
and an  (approximate) guarantee that only shallow backtracks will be needed
(in subproblems but not across subproblems).
% Furthermore,
The proposed approach has been integrated within the ROS environment,
and will be made available. 
%% long, convoluted sentence; avoid
%% in order to provide the highest level of standardization and compatibility with the maximum number of robotic systems currently available.
Nevertheless, there are still several interesting topics that could be treated in future work:
\begin{itemize}
    \item Inclusion of learning AI-based techniques to obtain self-learned rules, features and sampling strategies.

    \item Extension to consider optimality and being able to tackle problems with state uncertainty and partial observability, where dynamics- or physics-based planning is needed.

    \item Post-processing the obtained plans by using optimal motion-planning algorithms, e.g.~Informed~RRT$^*$~\cite{6942976}, and merging and shortening the trajectories of consecutive actions.
\end{itemize}

%% file: sections/acknowledgments.tex
\section{Acknowledgements}

This work is part of the action CNS2022-136178 financed by MCIN/AEI/10.13039/501100011033 and by the EU Next Generation EU/PRTR.
M.~Dalmau and N.~García have received funding from the EU’s Horizon Europe research and innovation programme under GA No. 101070136 - \mbox{INTELLIMAN}.
H.~Geffner has been supported by the Alexander von Humboldt Foundation with funds from the Federal Ministry for Education and Research; the European Research Council, GA~No.~885107; the EU's Horizon 2020 research and innovation programme, GA~No.~952215; the Excellence Strategy of the Federal Government and the NRW L\"{a}nder, Germany, and the Knut and Alice Wallenberg Foundation under the WASP program.

%% file: sections/background.tex
\section{Appendix: Supplementary Material}
This supplementary material contains a Background section and details of the extension of the available implementation of PDDLStream planners. Demonstration videos can be downloaded in~\url{https://anonymfile.com/bXOa/videos-ctmp-sketches.zip}
\subsection{Background}\label{sec:back}

We review  the notions of width and sketches developed in the setting of classical planning. 

\subsubsection{Classical Planning}
A \emph{classical planning} problem is a pair $P=\tup{\domain,\instance}$ where~$\domain$ is a first-order \emph{domain} with action schemas defined over predicates
and $\instance$ contains the objects in the instance and two sets of ground literals, the initial and goal situations $\initial$ and $\goal$ respectively
\cite{geffner:book,ghallab:book}. 
An instance $P$ defines a state model $S(P)=\tup{\states,\initialstate,\goalstates,\actions,\applicability(\cdot),\successor(\cdot,\cdot)}$ where
the states in $S$ are the truth valuations over the ground atoms represented by the set of literals that they make \texttt{true},
the initial state $s_0$ is $\initial$, the set of goal states $\goalstates$ are those that make the goal literals in $\goal$ \texttt{true},
and the actions $Act$ are the ground actions obtained from the schemas and objects. The ground actions in
$\applicability(s)$ are the ones that are applicable in a state~$s$; namely, those whose preconditions are \texttt{true} in~$s$,
and the state transition function $f$ maps a state $s$ and an action $a\! \in\! \applicability(s)$ into the successor state $s'\!=\!f(a,s)$.
A \emph{plan} $\pi$ for $P$ is a sequence of actions $a_0,\ldots,a_n$ that is executable in $s_0$ and maps the initial state $s_0$
into a goal state; i.e., $a_i\! \in\! \applicability(s_i)$, $s_{i+1}\!=\!f(a_i,s_i)$, and $s_{n+1}\!\in\!\goalstates$.
A state~$s$ is \emph{solvable} if a plan exists starting at $s$, otherwise it is \emph{unsolvable} (also called \emph{dead-end}).
The plan \emph{length} is the number of actions, and a plan is \emph{optimal} if  no shorter one exists.

\subsubsection{Width}

\iw{1} is a simple planning algorithm that makes use of a set of atoms or Boolean features $F$ to implement a breadth-first search
where a newly generated state is pruned if it does not make \texttt{true} an atom or feature in $F$ for the first time in the search \cite{lipovetzky-geffner-ecai2012}.
The procedure \iw{k} is like \iw{1} but prunes a state instead when it does not make \texttt{true}  a collection of up to $k$ atoms for the first time in the search.
Unlike breadth-first search, \iw{k} runs in time and space that are exponential in $k$ and not in the number of problem variables. 
It has been found effective in classical planning because \iw{k} for $k=1$ or $k=2$ is complete and optimal for many planning domains
when goals are single atoms. Serialized Iterated Width (SIW) uses \iw{k} with small values of $k$ for problems with 
conjunctive goals, that are tackled in sequence. \siw starts at the initial state~$s=s_0$ of $P$, and performs an \iw{k} search from
$s$ to find a shortest path to a state $s'$ such that $\#g(s') \!<\!\#g(s)$, where~$\#g(s)$ counts the number of unsatisfied top-level goals of
$P$ in~$s$. If~$s'$ is not a goal state, $s$ is set to $s'$ and the loop repeats.

These methods have been found effective in classical planning, because many classes of problems have a small \emph{width}, over which polynomial \iw{k} algorithms are very effective,
and that many other problems can be easily split into problems of small width~\cite{nir:aaai2017}. This is the decomposition exploited by \siw. 

The  width $w(P)$ of a planning problem $P$ is the minimum~$k$ for which there exists a  sequence
$t_0,t_1,\ldots,t_m$  of atom tuples $t_i$ from $P$,  each consisting of at most $k$ atoms, such that:
1)~$t_0$ is \texttt{true} in the initial state $\initialstate$ of $P$,
2)~any optimal plan for $t_i$ can be extended into an optimal plan for $t_{i+1}$ by adding a single action,
and 3)~if $\pi$ is an optimal plan for $t_m$, then  $\pi$ is an optimal plan for $P$.
The algorithm IW runs \iw{k} sequentially for $k=1,2, \ldots$ until the problem is solved or $k$
is the number of problem variables. IW solves a problem $P$ in time and space that are exponential
in $w(P)$. For convenience, when a problem $P$ is unsolvable, $w(P)$ is set to the number of variables in $P$,
and when $P$ is solvable in at most one step,  $w(P)=0$ \cite{lipovetzky-geffner-ecai2012,bonet-geffner-aaai2021}.
The width of a conjunction of atoms $T$ (or set of goal  states $S'$) is defined as the width of a
problem $P'$ that is like $P$ but with the goal $T$ (resp. $S'$).

An important feature of width-based planning is that, unlike other classical planning methods,
it does not require a declarative, PDDL-like, action model, and just requires a set of Boolean state features $F$.
This allows width-based planning  methods to be applied in domains for which a black-box simulator
computes the next states as in video games~\cite{nir:ijcai2015}.

\subsubsection{Sketches}

The \siw algorithm decomposes problems into subproblems which are solved by running IW.
The decomposition is done by exploiting the structure of conjunctive goals and requiring
to achieve one more top goal in each iteration. There are  problems however that
require more subtle forms of decompositions. The language of sketches
has been introduced recently for expressing such decompositions in
a convenient, compact way by means of simple rules over features.

A sketch $R$ is a collection of rules of the form $C \mapsto E$ where the condition $C$ and the effect
expression $E$ are defined over a set of state features $F$ that can be Boolean or numerical, 
taking values  over the non-negative integers \cite{bonet-geffner-aaai2021,drexler-et-al-kr2021}. 
The condition $C$ may contain expressions of the form $p$ or $\neg p$,  for a Boolean feature $p$ in $F$
or expressions  $n=0$ or $n>0$,  for a numerical feature $n$ in $F$.
Likewise, the effects $E$ may contain expressions of the form $p$, $\neg p$, and $\UNK{p}$ for
for a Boolean feature $p$ in $F$, and expressions of the form  $\DEC{n}$, $\INC{n}$, or $\UNK{n}$,
for a numerical feature $n$ in $F$. If $F=\{p,q,n,m\}$, the meaning of a rule $\prule{\neg p,n > 0}{p, \DEC{n}, \UNK{m}}$
is that in a state $s$ where $p$ is false and $n$ is positive, a \emph{subgoal state}~$s'$ needs to be found from $s$
where $p$ is \texttt{true}, $n$ decreases in value relative to its value in $s$, and $m$ can change arbitrarily.  Features like $q \in F$
that are not mentioned in the effects cannot change, and its value in $s$ and $s'$ must be the same. 

More precisely, a sketch~$R$ made of rules $r: C\!\mapsto\!E$ over a set of features $F$,
defines for each state $s$ of a problem $P$, a subproblem $P[s,G_R(s)]$ that is like $P$
but with initial state $s$ and a set of subgoal states  $G_R(s)$. This set is given 
by the states $s'$ such that the pairs  $[s,s']$ satisfies a  rule $C \!\mapsto\! E$ in $R$;
namely $s$ makes $C$ \texttt{true}, and the pair~$[s,s']$ makes the effect expression $E$ \texttt{true}\footnote{
  Usually $G_R(s)$ also contains the goal states of the problem $P$, but we do not need
  this extension in this work.}.
For instance, if $E\!=\!\{p,\DEC{n},\UNK{m}\}$ and the set of features is $F=\{p,q,n,m\}$,
then $[s,s']$ makes $E$ \texttt{true} if $p$ is \texttt{true} in $s'$, the value of $n$ in $s'$ is lower than the value of $n$ in $s$,
and the value of the feature~$q$, not mentioned in $E$, is the same in $s$ and $s'$.

The goal decomposition that underlies the \siwR\ algorithm is given by a sketch $R$
with a single rule $\prule{\#g > 0}{\DEC{\#g}}$, where $\#g$ is a feature
counting the number of unachieved top goals, and which defines
the set of subgoals $G_R(s)$ from a state $s$ as the states $s'$
when the number of unachieved top goals has been decreased.

\subsubsection{Sketch Width}

By decomposing problems, sketches can be used more flesibly with IW search methods than \siw. In particular,  \siwR\  uses a given sketch $R$ for solving
a class $\Q$ of problems $P$ in the following way. Starting  at the initial state $s=s_0$ of $P$,
an IW search is run to find a closest  state $s'$ in $G_R(s)$. If $s'$ is not a goal state of $P$,  $s$ is set to~$s'$,
and the process  repeats until a goal state is reached.
The width of a sketch $R$ over a class of problems $\Q$ is given by the maximum width over the subproblems
$P[s,G_R(s)]$ that may be encountered in the \siwR\ search. The \siwR\ algorithms thus solves
problems $P$ in time and space exponential in the sketch width. The use of
hand-made sketches in planning and ways for learning sketches of bounded width
 are considered in \cite{drexler-et-al-kr2021,drexler2022learning}.
For these complexity bounds, the features are assumed to be linear in $N$; namely,
they must have a linear number of values at most, computable in linear time \cite{bonet-geffner-aaai2021}.

%% file: sections/supplementary-pddlstream.tex
\subsection{PDDLStream Comparison Details}\label{sec:sup-pddlstream}
We compared our approach with PDDLStream~\cite{garrett_pddlstream_2020}, and it is important to remark that there exists a lack of work on cross-evaluation within the TAMP community. 
Although the selected benchmark~\cite{Lagriffoul2018PlatformIndependentBF} aims to be a standard for the community is not yet widely adopted. 
Thus, we decided to contribute also by evaluating an important work as PDDLStream in this benchmark.

PDDLStream is an AI planning framework with an action language and algorithms tailored for planning in scenarios involving sampling procedures. It enhances PDDL by introducing declarative stream specifications and solves problems without requiring detailed sampler descriptions. Its primary use case was in general-purpose robot TAMP.
This planning framework comes with four different algorithms: \emph{Incremental}, \emph{Focused}, \emph{Binding} and \emph{Adaptive}.

The \emph{Incremental} algorithm, iteratively increases the number of stream evaluations that are required to certify a fact. Furthermore, it eagerly and blindly evaluates all stream instances (required in a certain iteration) producing many facts that are irrelevant to the task. 
The other three algorithms belong to the category of Optimistic Algorithms, which means that they lazily explore candidate plans before checking their validity. 
Note that, in our work, we also propose lazy-evaluation of the geometric constrains until a plan candidate is found (refer to the main part of the paper for a complete description of this procedure).
\emph{Focused} is the starting point of the other two optimistic algorithms. It evaluates streams instances that have satisfied domain facts and adds new certified facts until a all plan actions are certified and a solution is found.
\emph{Binding} algorithm propagates stream outputs that are inputs to subsequent streams to evaluate more of the stream plan at once.
Finally, \emph{Adaptive} balances the time spent in searching versus processing the streams, often reducing the number of search calls required to find a solution.

We extended the available PDDLStream software\footnote{Find the repository in \url{https://github.com/caelan/pddlstream}.} to run the different algorithms in the multiple problem instances and variations of the benchmark~\cite{Lagriffoul2018PlatformIndependentBF}.
In particular, we modeled the different scenarios in the simulator Pybullet~\cite{coumans2021}. It is important to mention that we have not been able to incorporate the fence that surrounds the scenario in the defined benchmark in these experiments. Nevertheless, this benefits the PDDLStreams methods since with this modification the robot is less restricted and have more opportunities to interact with the environment objects. Thus, it has a little advantage particularly in cluttered environments.
Furthermore we defined and implemented the goal regions and conditions and solved multiple integration issues. 

In the Experimentation section, the evaluation results are reported. 
The more complex benchmark instances are not solved by any of the algorithms given the 30~min and 16~GB RAM budgets. 
In the case of the \emph{Incremental} algorithm, the timeout was exceeded and on the other cases the memory capacity was exceeded approximately when the 60\% of the time budget was reached.